\documentclass[runningheads]{llncs}
\usepackage{amsmath} %added by XZ
\usepackage{array}
\newcolumntype{P}[1]{>{\centering\arraybackslash}p{#1}}
\newcolumntype{M}[1]{>{\centering\arraybackslash}m{#1}}
\usepackage{amsfonts}%added by XZ
\usepackage{xcolor}%added by XZ
\usepackage{graphicx}

\usepackage[utf8]{inputenc}%added by XZ to have accent in the names like Barré ..

\begin{document}
\title{Towards Integrating Formal Verification of Autonomous Robots with Battery Prognostics and Health Management\thanks{Supported by the UK EPSRC through the Offshore Robotics for Certification of Assets (ORCA) [EP/R026173/1], Robotics and Artificial Intelligence for Nuclear (RAIN) [EP/R026084] and Science of Sensor System Software (S4) [EP/N007565].}}
\titlerunning{Formal Verification of Autonomous Robots and Battery PHM}
% If the paper title is too long for the running head, you can set
% an abbreviated paper title here
%
\author{
Xingyu Zhao\inst{1} \and
Matt Osborne\inst{1} \and
Jenny Lantair\inst{1} \and
Valentin Robu\inst{1}\and
David Flynn\inst{1}\and \\
Xiaowei Huang\inst{2}\and
Michael Fisher\inst{2}\and
Fabio Papacchini\inst{2}\and
Angelo Ferrando\inst{2}}
\authorrunning{X. Zhao et al.}
% First names are abbreviated in the running head.
% If there are more than two authors, 'et al.' is used.
%
\institute{
School of Engineering and Physical Sciences,\\ Heriot-Watt University, Edinburgh EH14 4AS, U.K.\\
\email{\{xingyu.zhao,mho1,jl153,v.robu,d.flynn\}@hw.ac.uk}\\
\and
Department of Computer Science, \\
University of Liverpool, Liverpool L69 3BX, U.K.
\email{\{xiaowei.huang,mfisher,fabio.papacchini,angelo.ferrando\}@liverpool.ac.uk}}
\maketitle              % typeset the header of the contribution
\begin{abstract}
The battery is a key component of autonomous robots. Its performance limits the robot's safety and reliability. Unlike liquid-fuel, a battery, as a chemical device, exhibits complicated features, including (i) capacity fade over successive recharges and (ii) increasing discharge rate as the state of charge (SOC) goes down for a given power demand. Existing formal verification studies of autonomous robots, when considering energy constraints, formalise the energy component in a generic manner such that the battery features are overlooked. In this paper, we model an unmanned aerial vehicle (UAV) inspection mission on a wind farm and via probabilistic model checking in PRISM show (i) how the battery features may affect the verification results significantly in practical cases; and (ii) how the battery features, together with dynamic environments and battery safety strategies, jointly affect the verification results. Potential solutions to explicitly integrate battery prognostics and health management (PHM) with formal verification of autonomous robots are also discussed to motivate future work.

\end{abstract}
\section{Introduction}

Autonomous robots, such as unmanned aerial vehicles (UAV) (commonly termed drones\footnote{We have used the word ``drone'' interchangeably with the abbreviation UAV as a less formal naming convention throughout the paper.}), unmanned underwater vehicles (UUV), self-driving cars and legged-robots, obtain increasingly widespread applications in many domains \cite{guiochet_safety_2017}. Extreme environments -- a term used by UK EPSRC\footnote{https:\//epsrc.ukri.org/files/funding/calls/2017/raihubs} to denote environments that are remote and hazardous for humans --  are the most promising domains in which autonomous robots can be deployed to carry out a task, such as exploration, inspection of oil/gas equipment on the seabed, maintenance of offshore wind turbines, and  monitoring of nuclear plants in high radiation conditions \cite{lane_new_2016}.

However, autonomy poses a great challenge to the assurance of safety and reliability of robots, whose failures may cause both a detriment to human health and well-being and huge financial losses. Thus, there are increasing demands on regulation of autonomous robots to build public trust in their use, whilst the development, verification and certification of autonomous robots is inherently difficult due to the sheer complexity of the system design and inevitable uncertainties in their operation \cite{fisher_verifying_2013,farrell_robotics_2018,fisher_verifiable_2018,robu_train_2018}. For instance, \cite{kalra_driving_2016} shows the infeasibility of demonstrating the safety of self-driving cars from road testing alone, and both \cite{koopman_autonomous_2017} and \cite{kalra_driving_2016} argue the need for alternative verification methods to supplement testing. Formal techniques, e.g. model checking and theorem proving, offer a substantial opportunity in this direction \cite{giaquinta_strategy_2018}. Indeed, formal methods for autonomous robots have received great attention \cite{farrell_robotics_2018,luckcuck2018formal}, both in controller synthesis, see e.g. \cite{giaquinta_strategy_2018,norman_verification_2017}, and in verifying safety and reliability when the control policy is given, see e.g. \cite{zhao_probabilistic_2019,hoffmann_autonomous_2016,gerasimou_efficient_2014}.

The battery as the power source of autonomous robots plays a key role in real-life missions \cite{zhang_estimation_2012}. However to the best of our knowledge, most existing formal verification studies of autonomous robots, when considering energy constraints, formalise the energy component in a generic and simplified manner such that some battery features are overlooked:
\begin{itemize}
    \item \textbf{Capacity fading}: Over successive recharges, unlike a liquid-fuelled system whose tank volume normally remains unchanged, the charge storage capacity of a battery will diminish over time.
    \item \textbf{Increasing discharge rate}: In one discharge cycle, since the voltage drops as the battery is being discharged, for a constant power output (a product of the voltage and the current), the current increases meaning an increased discharge rate occurs. This is different to a liquid-fuelled system in which a constant power output typically means a constant rate of fuel consumption.
\end{itemize} 
Thus, usual assumptions, like (i) a fixed battery capacity regardless the number of recharges and (ii) constant energy consumption for a given action regardless the stage in a discharge cycle, become potentially problematic. 

On the other hand, the battery prognostics and health management (PHM) community has been developing techniques to accurately forecast the battery behaviour in both a life-cycle and a discharge-cycle. We believe such battery PHM results should be integrated into formal studies (either controller synthesis or verification) of robots to refine the analysis. To take a step forward in this direction, in this paper, our main work is as follows:
\begin{itemize}
    \item We formalise a UAV inspection mission on an offshore wind farm, in which the mission scenario and choice of model parameters are based on a real industry survey project. The UAV takes a sequence of actions and follows a fixed inspection route on a $6\times 6$ wind farm. It autonomously decides when to return to the base for recharges based on the health/states of the battery. Uncertainties come from the dynamic environment which causes different levels of power demand.  %\xiaowei{a brief description on the model?} ok. done.
    \item We explicitly consider the two battery features in our modelling and show (i) how different battery safety strategies, dynamic environments (i.e. different levels of power demand) and the battery chemical features jointly affect the formal verification results; and (ii) the verification results could be either dangerously optimistic or too pessimistic in practical cases, without the modelling of the battery features. %\xiaowei{would be useful to discuss more on the experimental results.}
    \item We discuss important future work on explicitly integrating battery PHM with formal verification, given the trend that advanced PHM algorithms are mostly based on real-time %\xiaowei{change up-to-date  to real-time?} done.
    readings from sensors deployed on the battery.
\end{itemize}

The organisation of the paper is as follows. In the next section, we present preliminaries on probabilistic model checking and battery PHM. The running example is described in Sec.~\ref{sec_example}. We show our probabilistic model and verification results in Sec.~\ref{sec_model_PRISM} and \ref{sec_results}, respectively. Sec.~\ref{sec_related_work} summarises the related work. Future work and contributions are concluded in Sec.~\ref{sec_conclusions}.

\section{Background}
\label{sec_background}

\subsection{Probabilistic Model Checking}
Probabilistic model checking (PMC) \cite{kwiatkowska_probabilistic_2018} has been successfully used to analyse quantitative properties of systems across a variety of application domains, including robotics \cite{luckcuck2018formal}. This involves the construction of a probabilistic model, commonly using Discrete Time Markov Chain (DTMC), Continuous Time Markov Chain (CTMC) or Markov Decision Process (MDP), that formally represent the behaviour of a system over time. The properties of interest are usually specified with e.g., Linear Temporal Logic (LTL) or Probabilistic Computational Tree Logic (PCTL), and then systematic exploration and analysis is performed to check if a claimed property holds. In this paper, we adopt DTMC and PCTL whose definitions are as follows. 

\textbf{Definition 1}. A DTMC is a tuple $(S,s_1,\textbf{P},L)$, where:
\begin{itemize}
\item $S$ is a (finite) set of states; and $s_1\in S$ is an initial state;
\item $\textbf{P}:S\times S \rightarrow [0,1]$ is a probabilistic transition matrix such that $\sum_{s^{\prime}\in S}\textbf{P}(s,s^\prime)=1$ for all $s\in S$;
\item $L:S\rightarrow 2^{AP}$ is a labelling function assigning to each state a set of atomic propositions from a set $AP$.
\end{itemize}

\textbf{Definition 2}. $AP$ is a set of atomic propositions and $ap \in AP, p\in [0,1], t\in \mathbb{N}$ and $\bowtie \in \{<,\leq,>,\geq\}$. The syntax of PCTL is defined by \textit{state formulae} $\Phi$ and \textit{path formulae} $\Psi$.
\begin{align}
\Phi &::= true \mid ap \mid \Phi \wedge \Phi \mid \neg \Phi \mid \mathcal{P}_{\bowtie p}(\Psi) \nonumber
\\
\Psi &::= X \: \Phi \mid \Phi \: U^{\leq t} \: \Phi \mid \Phi \: U \: \Phi \nonumber  
\end{align}
where the temporal operator $X$ is called ``next'', $U^{\leq t}$ is called ``bounded until'' and $U$ is called ``until''. Also, $F \: \Phi$ is normally defined as $true \: U \: \Phi $ which is called ``eventually''.
State formulae $\Phi$ is evaluated to be either true or false in each state. Satisfaction relations for a state $s$ are defined:
\begin{align}
s & \models true \nonumber \quad,\quad
s \models ap \quad\text{iff}\quad ap \in L(s)  \nonumber
\\
s & \models \neg \Phi \quad\text{iff}\quad s \not\models \Phi  \nonumber
\\
s & \models \Phi_1\wedge\Phi_2 \quad\text{iff}\quad s \models \Phi_1 \text{ and } s \models \Phi_2  \nonumber
\\
s & \models \mathcal{P}_{\bowtie p}(\Psi) \quad\text{iff}\quad Pr(s\models \Psi)\bowtie p \nonumber
\end{align}
$Pr(s\models \Psi)\bowtie p $ is the probability of the set of paths starting in $s$ and satisfying $\Psi$. Given a path $\psi$, if denote its \textit{i}-th state as $\psi[i]$ and $\psi[0]$ is the initial state. Then the satisfaction relation for a path formula for a path $\psi$ is defined as:
\begin{align}
\psi & \models X \Phi \quad\text{iff}\quad \psi[1] \models \Phi \nonumber
\\
\psi & \models \Phi_1 U^{\leq t}\Phi_2 \quad\text{iff}\quad \exists 0 \leq j \leq t \nonumber
\\
& (\psi[j]\models \Phi_2\wedge(\forall 0\leq k<j \; \psi[k]\models \Phi_1)) \nonumber
\end{align}

It is worthwhile mentioning that both DTMC and PCTL can be augmented with rewards/costs \cite{filieri_probabilistic_2013}, which can be used to model, e.g. the energy consumption of robots in a mission. Indeed, this is the typical way used in existing studies, and differs from our modelling of battery in this study.

After formalising the system and its requirements in DTMC and PCTL, respectively, the verification focus shifts to the checking of \textit{reachability} in a DTMC. In other words, PCTL expresses the constraints that must be satisfied, concerning the probability of, starting from the initial state, reaching some states labelled as, e.g. unsafe, success, etc. Automated tools have been developed to solve the reachability problem. We use PRISM \cite{kwiatkowska_prism_2011} which employs a symbolic model checking algorithm to calculate the probability that a path formulae is satisfied. More often, it is of interest to know the actual probability that a path formula is satisfied, rather than just whether or not the probability meets a required bound. So the PCTL definition can be extended to allow \textit{numerical queries} by the form $\mathcal{P}_{=?}(\Psi)$ \cite{kwiatkowska_probabilistic_2018}.

In general, a PRISM module contains a number of local variables which constitute the state of the module. The transition behaviour of the states in a module is described by a set of commands which take the form of:
\begin{equation}
    [Action] \: Guard \rightarrow Prob_1 : Update_1 + ... + Prob_n : Update_n ; \nonumber
\end{equation}
As described by the PRISM manual\footnote{https://www.prismmodelchecker.org/manual/}, the guard is a predicate over all the variables (including those belonging to other modules. Thus, together with the action labels, it allows modules to synchronise). Each update describes a transition which the module can make if the guard is true. A transition is specified by giving the new values of the variables in the module, possibly as a function of other variables. Each update is also assigned a probability (in our DTMC case) which will be assigned to the corresponding transition.

\subsection{Battery Modelling and PHM}

\begin{figure}[t]
	\centering
	\includegraphics[width=1.0\textwidth]{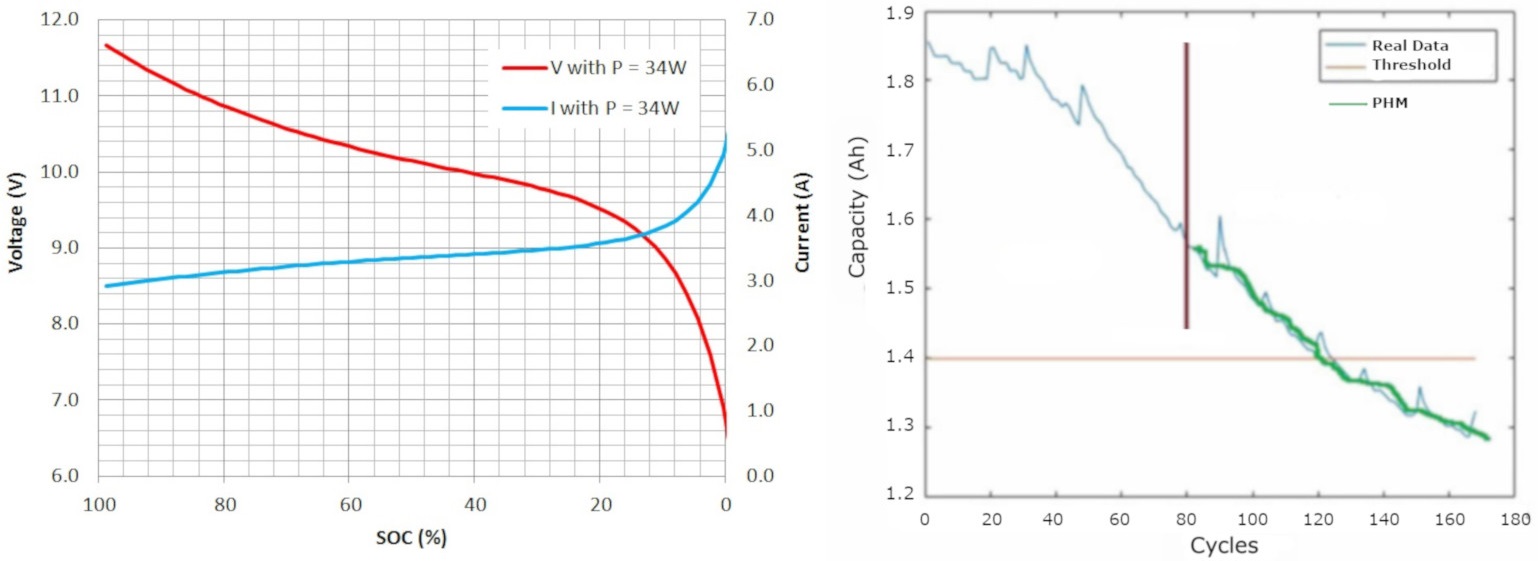}
	\caption{(Left) The non-linear dynamics of voltage and current vs SOC for a constant power demand from \cite{traub_calculation_2016}. (Right) Cited from \cite{andoni_data_2017}, the ``real data'' curve showing a Lithium-ion battery capacity fade and its PHM predictions (thick green line).}
	\label{constPower_time}
\end{figure}

Electric batteries exhibit non-linear charge and discharge characteristics due to a number of factors. The voltage varies with the state of charge (SOC) because of changing chemical properties within the cell, such as increasing electrolyte resistance, non-linear diffusion dynamics and Warburg inductance \cite{HariharanKrishnanS2018MMoL}. Fig.~\ref{constPower_time}-(Left), derived from the experimental test in \cite{traub_calculation_2016}, shows such non-linear results of voltage and current vs SOC profile for a constant power demand.

%Here an equation has been used to transform the voltage curve from a constant-current voltage curve and the corresponding constant power current curve is derived.

%The voltage is also affected by the cell temperature and must be kept within safe operating limits. These complex battery model characteristics are not considered in this paper but can be introduced in future Prognostics battery models.

%In order to deliver a constant power output with a falling voltage, a discharge curve must be calculated from a constant current discharge experimental test. 

A constant power demand means that an increase in current is drawn as the voltage falls with SOC. For our study, we are interested in a UAV with a battery capacity of around 11Ah and nominal voltage of 22V. The energy supply is 180Wh from a lithium polymer battery. For a 22V battery the voltage at full charge is $\sim$25V and will drop to $\sim$20V at a safe threshold of 30\%  SOC.

The easiest way to measure a change in SOC is by integrating the current discharge over time from a known initial SOC, called Coulomb counting \cite{he2013state}:
\begin{equation}
\label{eq_coulomb_counting}
  SOC(k+1) = SOC(k) - \frac{I(k) \times \Delta t}{Q_{max}}
\end{equation}
where $Q_{max}$ is the maximum SOC, $I(k)$ is the time dependant current, $SOC(k)$ is the SOC percentage at the discrete time step $k$, $\Delta t$ is the time step interval. Although this simplification does not take into consideration inaccuracies in the battery initial SOC estimation or account for the internal losses, it is proposed as a first approximation to model the power usage and discharge characteristics as discrete states using the known battery characteristics.

Batteries also degrade over successive recharges due to decreased lithium-ion concentrations so that over 1000 discharge cycles a 20\% loss in capacity may occur \cite{spotnitz_simulation_2003}. Fig.~\ref{constPower_time}-(Right) shows a typical lithium-ion battery capacity fade characteristics the (``real data'' curve), cited from \cite{andoni_data_2017}. We can observe, after the first 80 discharge cycles, the battery's capacity drops from 1.85(Ah) to 1.55(Ah).

%\begin{figure}[ht]
%	\centering
%	\includegraphics[width=0.8\textwidth]{fig_capacity_fade_PHM.jpg}
%	\caption{Cited from \cite{andoni_data_2017}, ``real data'' curve shows Lithium-ion battery capacity fading. And the PHM predictions of the battery's capacity (thick green line).}
%	\label{batteryFade}
%\end{figure}
%}

There is a growing interest in the use of PHM techniques to reduce life-cycle costs for complex systems and core infrastructure \cite{Goebel2017}. Battery health management is also a critical area in regards to the safe and reliable deployment of UAVs. Numerous studies into battery PHM techniques have been carried out, e.g. the use of Neural Nets \cite{gao2017novel}, Unscented Kalman Filters \cite{he2013state,hogge_verification_2018}, Unscented Transform \cite{daigle2010improving}, Hardy Space $H_\infty$ Observers \cite{zhang_estimation_2012} and Physics Based models \cite{he2018physics}. Although we are assuming a hypothetical/generic battery PHM method in this paper to provide parameters in our latter modelling, it is envisaged that advanced PHM techniques can be integrated in our future verification framework.

\section{The Running Example}
\label{sec_example}

As UAV technology improves, energy companies are looking to adopt the technology to reduce maintenance and operating costs. The \textit{resident drone} idea is to station a UAV at locations where aerial surveys are conducted repeatedly. The advantages of such resident drone inspection system are the possibility of increased availability for data collection (e.g. to feed in techniques reviewed in \cite{stetco_machine_2019}), reduced manual labour, improved safety and more cost effective maintenance strategies. We model a typical application of such system in this paper, based on a survey utilising commercial technologies.

%This paper introduces a model for an off-shore wind farm resident drone inspection mission. Current commercial technology for drones and lithium polymer battery systems are used to derive the battery and drone performance.

\begin{figure}[ht]
	\centering
	\includegraphics[width=1\textwidth]{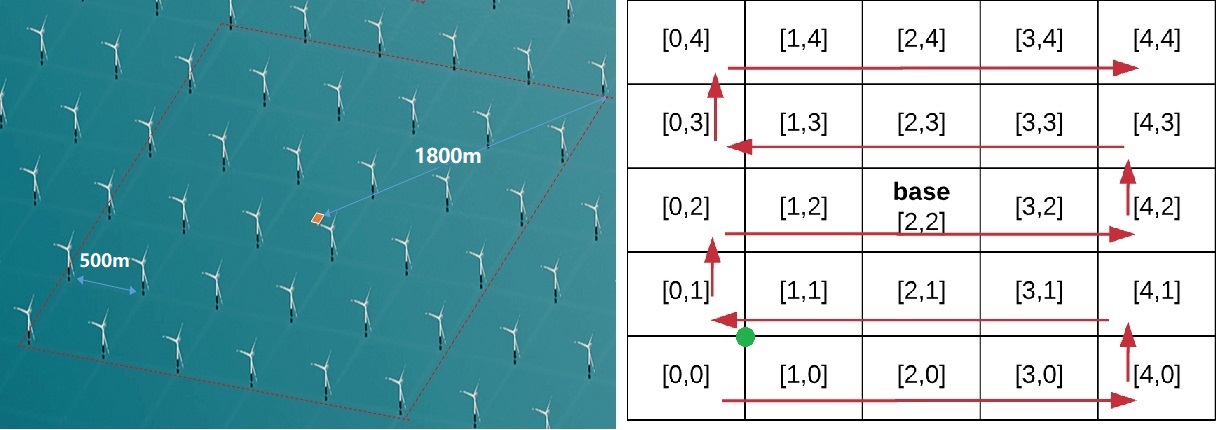}
	\caption{(Left) A fully autonomous UAV inspection mission in a $6\times6$ wind farm. (Right) The fixed controller of a UAV inspection mission on the wind farm. Intersections and cell spaces represent wind turbines and transportation channels respectively.}
	\label{fig_wind_farm}
\end{figure}

A simplified wind farm drone inspection mission as a $6 \times 6$ grid of turbines with a UAV located at the centre is considered. Wind turbines are typically distributed between 5 and 12 turbine blade diameters apart, 
%a greater distance in the direction of the wind to reduce wake interference \cite{ng2016offshore}. 
so a square distribution of turbines is modelled, each 500 meters apart, as shown in Fig.~\ref{fig_wind_farm}-(Left).

The drone mission requires the drone take-off and land at the base station, fly a distance determined by the number of grid spaces to a turbine, carry out an inspection and return to the base or continue the mission with a single battery charge. For this mission a drone transit velocity of $\sim$10 m/s is assumed. An inspection is expected to take 15 minutes and the take-off and landing time is estimated less than 1 minute. The battery recharge time is around 1.5 hours.% I changed here, basically, 1 min for takeoff/landing. 2 minutes per cell travel.

\section {The Modelling in PRISM}
\label{sec_model_PRISM}

Our formal model of the running example presented in section \ref{sec_example} is a product (via parallel composition in PRISM) of four modules  -- \textit{Drone}, \textit{Grid}, \textit{Environment} and \textit{Battery}. Depending upon the model parameters used, a typical instance of our model has roughly $100,000$ states and $170,000$ transitions. In what follows, we introduce the modules separately and describe key assumptions, constants, and variables used in each module. Given the page limits, we only show some typical PRISM commands in the modules and omit some sophisticated synchronisation and parallel composition among modules. The complete sources code in the PRISM language can be found in our repository\footnote{https://x-y-zhao.github.io/files/VeriBatterySEFM19.prism.}.

\subsection{The \textit{Drone} Module}
\begin{figure}[ht]
\centering
\includegraphics[width=1\textwidth]{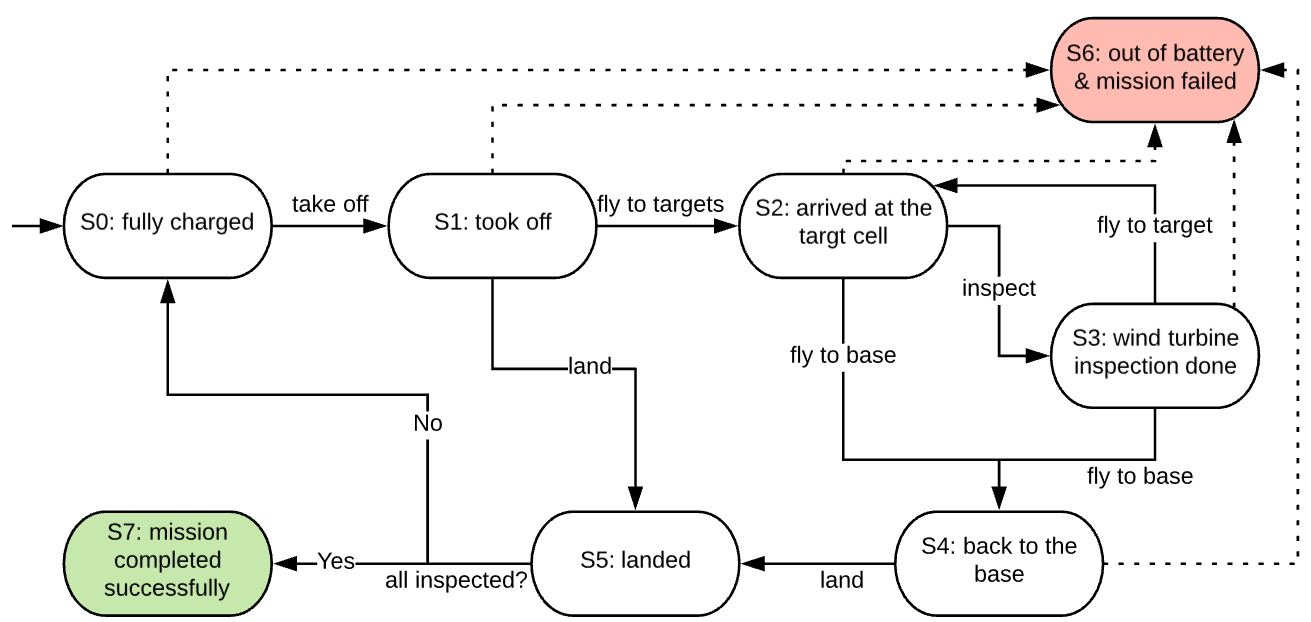}
\caption{A finite state machine of the UAV behaviour modelled by the \textit{Drone} module. }
\label{fig_state_mach_drone_module}
\end{figure}

The \textit{Drone} module is essentially a finite state machine describing the behaviour of the UAV during the inspection mission, as shown in Fig.~\ref{fig_state_mach_drone_module}. The UAV begins the mission in a fully charged state (S0) at the base. Once the UAV successfully takes off (S1), it may either directly land due to violation of the \textit{battery safety strategy} (see section \ref{sec_battery_module}), or fly to the target cell (S2) and then carry out an inspection of the wind turbine (S3). Depending upon the battery safety strategy and the battery SOC left after the inspection, the UAV will either fly back to the base for recharging (S4), stay in the same cell if there are more than one wind turbines to be inspected, or fly to the next target cell (S2) if all wind turbines of the current cell have been inspected. Once landed at the base (S5), the UAV will declare success of the mission if all wind turbines on the wind farm have been inspected, or recharge and continue the above work-flow otherwise. 

The dotted lines in Fig.~\ref{fig_state_mach_drone_module} represent events where the battery SOC falls to 0, leading to an out-of-battery state (S6). Note, the transition from S0 to S6 means that the fully charged capacity is not sufficient to do the next inspection at the target cell and thus the UAV declares the mission failed without further actions.

It is worthwhile to mention that, realistically, there should be some probability of failure for each action, e.g. $10^{-4}$ for landing. However, since we are only interested in the particular failure mode of out-of-battery here, we simplify our model by setting the failure probability of each action to 0. Thus, the only source of uncertainty we consider is from the dynamic environment which causes different levels of power demand for each action. We will discuss this in Sec.~\ref{sec:env}.

\subsection{The \textit{Grid} Module}

We formalise the wind farm as a $5\times 5$ grid as shown in Fig.~\ref{fig_wind_farm}-(Right) in which the intersections represent wind turbines and the cell spaces (labelled by coordinates $[x,y]$) represent transport channels. In this study, we assume a given control policy (CP) of the UAV as follows:
\begin{itemize}
    \item CP1: The UAV will follow the snake shaped route, as shown by the red arrows in Fig.~\ref{fig_wind_farm}-(Right), to carry out the inspection in each cell. 
    \item CP2: Depending upon the coordinates of the cell, there can be 1, 2 or 4 \textit{appointed} wind turbines to be inspected within a cell. For instance, at cell [0,0], the wind turbine located at the left-bottom corner is the only appointed one; both the two bottom corners at cell [2,0] need to be inspected; and for cell [2,2], all 4 wind turbines around it should be inspected. Indeed, it would be unwise (i.e. requiring more energy) to fly to cell [0,0] to inspect the green dotted wind turbine in Fig.~\ref{fig_wind_farm}-(Right), rather than fly with the shortest route to cell [1,1].
    \item CP3: Depending upon the battery safety strategy and the remaining SOC, the UAV may suspend the mission and return to the base for recharging. It will resume the mission at the cell where the mission was suspended.
\end{itemize}

\begin{figure}[ht]
\centering
\includegraphics[width=1\textwidth]{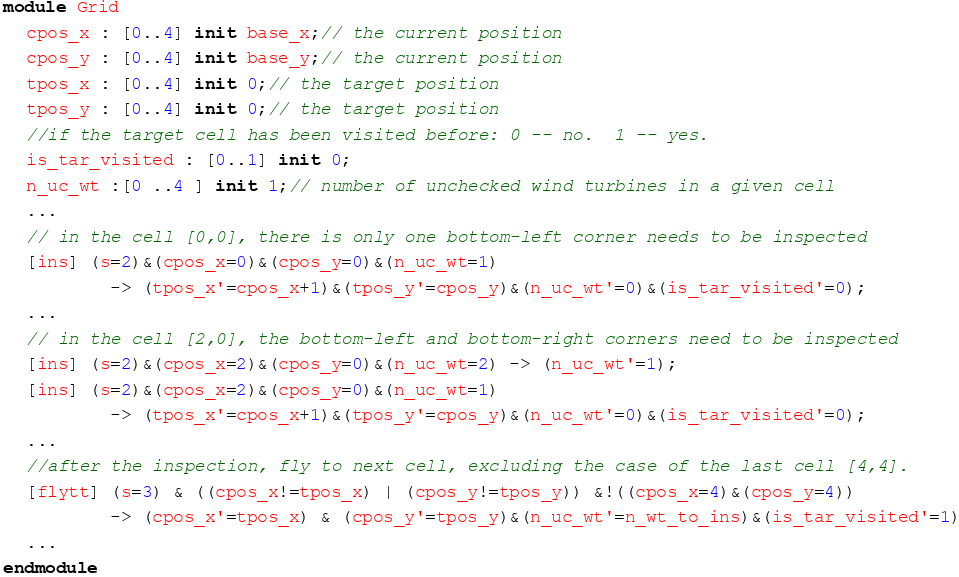}
\caption{Some PRISM commands of the \textit{Grid} module.}
\label{fig_PRISM_commands_GRID}
\end{figure}

A part of the PRISM commands in this module are shown in Fig.~\ref{fig_PRISM_commands_GRID}. Note, the transitions probabilities are simplified to 1.

\subsection{The \textit{Environment} Module}\label{sec:env}

We explicitly consider the environmental dynamics due to its primary impact on the battery's power demand. For simplicity, only one major factor -- wind speed -- was considered when developing the \textit{Environment} module. We formalise two levels of wind speed, and use a parameter $\mathit{p\_{wsp}\_c}$ to capture the dynamics of the wind. Environmental assumptions (EA) are listed below:
\begin{itemize}
    \item EA1: The UAV will only attempt to take off in a low wind speed condition.
    \item EA2: The change of wind speed (either from low to high or the other way around) occurs, with a probability of $p\_{wsp}\_c$, before each action is taken in the \textit{Drone} module. 
\end{itemize}

\begin{figure}[ht]
\centering
\includegraphics[width=0.8\textwidth]{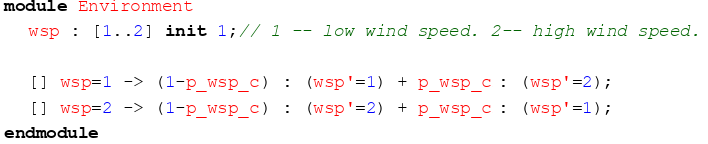}
\caption{The PRISM commands of the \textit{Environment} module.}
\label{fig_PRISM_commands_ENV}
\end{figure}

From EA2, we know that the higher the $p\_{wsp}\_c$ is, the more dynamic the environment, and it is this assumption that introduces uncertainty in the energy consumption for a given action. The PRISM commands are shown in Fig.~\ref{fig_PRISM_commands_ENV}.

\subsection {The \textit{Battery} Module}
\label{sec_battery_module}

\begin{figure}[ht]
\centering
\includegraphics[width=1\textwidth]{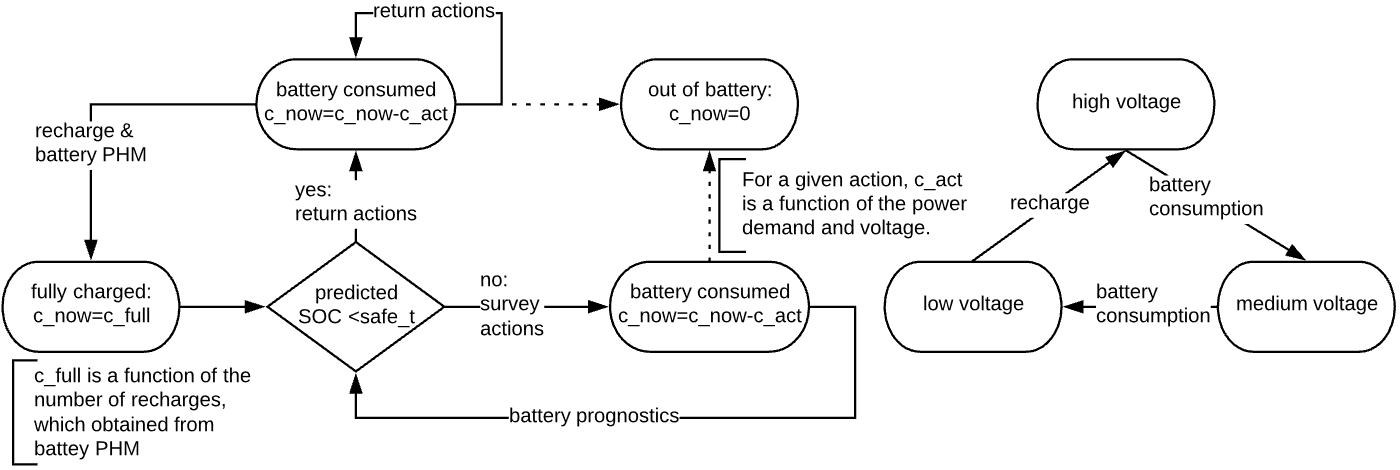}
\caption{Abstracted state machines of the \textit{Battery} module.}
\label{fig_battery_module}
\end{figure}

Fig.~\ref{fig_battery_module} shows two abstracted state machines of the \textit{Battery} module which run in parallel to describe the battery behaviour. The battery features of capacity fading (over successive recharges) and increasing discharge rate (in a single discharge cycle) are captured by battery assumptions (BA) as follows:
\begin{itemize}
    \item BA1: After each recharge, the battery's fully charged capacity (i.e. $c\_{full}$ in Fig.~\ref{fig_battery_module}) cannot be recovered to the new battery's capacity. 
    Rather, it decreases with a rate which can be obtained from battery PHM experiments, e.g. as observed from the results in \cite{andoni_data_2017}, for the first 100-ish discharge cycles, at each recharge, the capacity will fade at an average rate of 0.2\%.
    \item BA2: For a given wind speed, the UAV is working at a constant power demand for all actions. Since we considered 2 levels of wind speed in the \textit{Environment} module, there are 2 levels of constant power demand as well.
    \item BA3: In one discharge cycle, the battery's voltage is essentially a non-linear function of its SOC. We use a step-wise function to approximate the non-linear function -- high voltage $V_2$ (SOC $>$ 0.75), medium voltage $V_1$ (0.75 $\geq$ SOC $\geq$ 0.25) and low voltage $V_0$ (SOC $<$ 0.25).
\end{itemize}

In line with the BA2 and BA3, we use the following Eq.~\eqref{eq_battery_consumption} to estimate the battery consumption (Ah) for an action $j$ (denoted as $c\_act$ in Fig.~\ref{fig_battery_module}) under different levels of voltage $V_{i}$ and a power demand level $k$:
\begin{equation}
\label{eq_battery_consumption}
   C_j=\frac{E_{spec}\cdot {t_j}} {{V_i} \cdot T_{k}}
\end{equation}
where $E_{spec}$ is the specified battery energy, $T_k$ is the total running time under a constant power level $k$, $V_i$ is the level of voltage, and $t_j$ is the estimated execution time of action $j$.

For instance, a typical UAV battery with a specified energy of 180Wh ($E_{spec}=180$) can fly 30 minutes at the normal level of workload ($T_k=0.5$ and $k$ represents the normal level of power demand). The specified normal working voltage is 22V ($V_{1}=22$) (with a maximum level of 25V, $V_{2}=25$, and minimum level of 20V, $V_{0}=20$). The average time for inspecting a wind turbine is 15 minutes (if action $j$ represents the inspection, then $t_j=0.25$). Via Eq.~\eqref{eq_battery_consumption} and those estimated parameters above, we obtain the last row in Table~\ref{tab_battery_capacity_use_actions_diff_voltage} (results are rounded to one decimal place). Similarly for the battery consumption of each action at the high power demand level (high wind speed environment), the values can be calculated in the same way but are not shown in this paper.

\begin{table}
\caption{Battery consumption (Ah) of actions under different levels of voltage in low wind speed environment (i.e. the normal level of power demand).}
\centering
\label{tab_battery_capacity_use_actions_diff_voltage}
\begin{tabular}{M{2.5cm}|c|c|c}
\hline
 & low voltage &  medium voltage & high voltage  \\
\hline
\hline
take-off/land & 0.3 & 0.2 & 0.1\\\hline
transport per cell & 0.5 & 0.4 & 0.3\\\hline
inspection per wind turbine & 4.5 & 4.0 & 3.6\\
\hline
\end{tabular}
\end{table}

So far, in our modelling, instead of assuming a fixed battery consumption for each action, we have 6 possibilities of the battery consumption after an action (3 voltage levels $\times$ 2 power demand levels).

Engineers are aware of the higher risks associated from operating with a lower SOC battery, thus there are requirements on the battery PHM to provide warnings when the SOC falls below a certain threshold \cite{saxena_requirements_2012} (and to recommend that the mission be discontinued), e.g. a typical 30\% threshold is adopted by NASA in \cite{hogge_verification_2018}. In line with that battery safety strategy (BS), we also define a parameter $\mathit{safe\_t}$ as a safety threshold:
\begin{itemize}
    \item BS1: before each of the actions, take-off, fly-to-target and inspect, the UAV will check if the SOC will fall below $\mathit{safe\_t}$ after a sequence of actions to perform an intended inspection. If there is sufficient SOC, then take the action, otherwise return for recharging.
\end{itemize}

\begin{figure}[ht]
\centering
\includegraphics[width=1\textwidth]{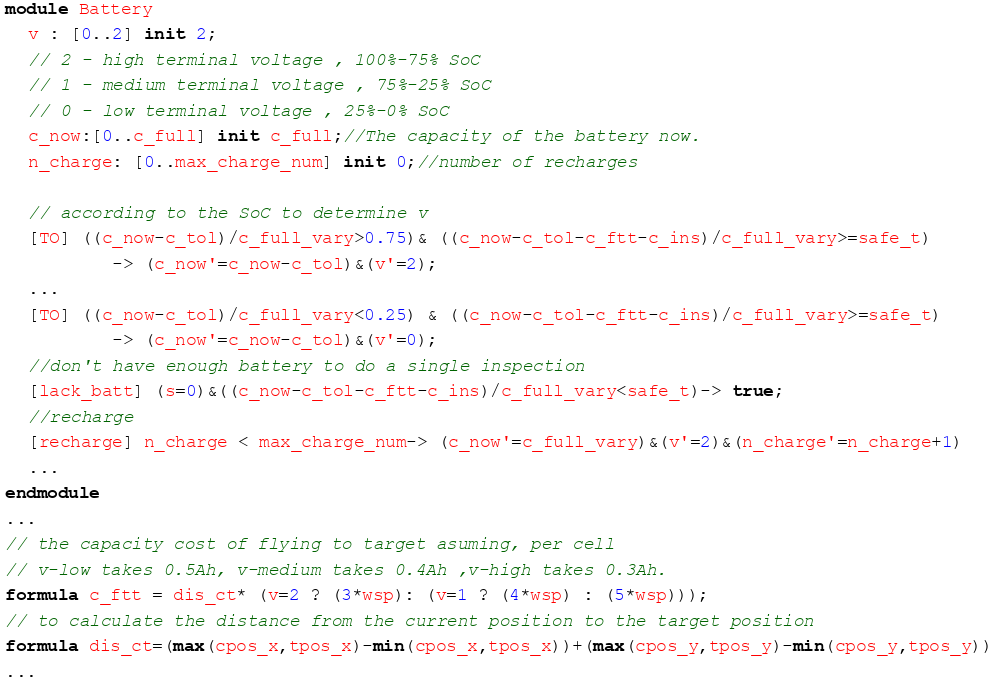}
\caption{Some PRISM commands of the \textit{Battery} module and global formulas. Note, the key variables $\mathit{c\_full\_vary}$ (the fully-charged capacity considering capacity fading) and $\mathit{c\_ftt}$, $\mathit{c\_ins}$ etc. (the battery consumption of each action, which are generically denoted as $c\_act$ in Fig.~\ref{fig_battery_module}) should be obtained dynamically from the PHM system in reality, whilst we make simplified assumptions in the source codes.
}
\label{fig_PRISM_commands_Battery}
\end{figure}

An instance of BS1 is that, before flying to the target cell, the UAV will predict the remaining SOC (based on the current battery health/state and wind conditions) \textit{after} flying to the target and performing one inspection. If there is no safe battery life remained (i.e. SOC$<\mathit{safe\_t}$) after the intended inspection, the UAV will go back for recharging. Some typical PRISM commands of this module and associated formulas are shown in Fig.~\ref{fig_PRISM_commands_Battery}.

%\xiaowei{can be useful to have an example explaining the global state (with local states from the four components), and a few example transitions. }

\section{Results}
\label{sec_results}

The main properties of interest and their corresponding PCTL formulas are: 
\begin{itemize}
    \item The probability of mission success\footnote{Since we focus on the particular failure mode of out-off-battery in our model, rigorously this should be the probability of seeing no out-off-battery failures in a mission.}: $ P_{=?} [ \; F \: (s=7) ]$;
    \item The expected mission time: $R\{\text{``mt''}\}_{=?} [\; F \; (s=7)|(s=6) ]$;
    \item The expected number of recharges: $R\{\text{``rc''}\}_{=?} [\; F \; (s=7)|(s=6) ]$.
\end{itemize}

We use the PRISM tool \cite{kwiatkowska_prism_2011} to check the properties given different model parameters in later subsections. Indeed, we may only be concerned with the expected mission time (or number of recharges) \textit{given} the mission is successful. However, PRISM can only solve the ``reachability reward'' properties when the target set of states is reached with probability 1, thus our target state here is $(s=7)|(s=6)$. In our later numerical examples, we only show the expected mission time when the probability of the mission failing is very small so that its contribution to the average mission time is negligible. Note, this limitation of PRISM has been studied in \cite{marcker_computing_2017}.

\subsection{Effects of Battery Safety Strategies and Dynamic Environments}

For a typical new battery with 11Ah capacity, we highlight the verification results of four representative cases, as shown in Table~\ref{tab_veri_typical_cases}, by setting the above mentioned model parameters (cf. BS1 and EA2) as:

\begin{itemize}
    \item \#1, the common case and baseline:  $\mathit{safe\_t}=0.3$, $\mathit{p\_wsp\_c}=0.1$.
    \item \#2, a risky battery strategy: $\mathit{safe\_t}=0.25$, $\mathit{p\_wsp\_c}=0.1$
    \item \#3, a more dynamic environment: $\mathit{safe\_t}=0.3$, $\mathit{p\_wsp\_c}=0.3$.
    \item \#4, a risky battery strategy in a more dynamic environment: $\mathit{safe\_t}=0.25$, $\mathit{p\_wsp\_c}=0.3$.
\end{itemize}

\begin{table}[htbp]
	\caption{Verification results of some typical cases with a new battery capacity of 11Ah.}
	\centering
	\label{tab_veri_typical_cases} 
	\begin{tabular}{c|c|M{1.7cm}|M{2cm}|M{1.9cm}|M{1.8cm}}
		\hline 
		& No. of states & No. of transitions& Prob. mission success & Exp. mission time& Exp. no. of recharges \\
		\hline \hline
    \#1	&108,688 &163,076 & 1 &4700.20 &42.65\\\hline
   \#2 &117,765 &177,278 &0.91 &3885.95 &34.59\\\hline
    \#3 &108,688 & 163,076 & 1 & 7482.86&72.16\\ \hline
    \#4 & 117,765 &177,278 & 0.89 & 5621.66 &53.07\\
		\hline
	\end{tabular}
\end{table}

The example of \#1 represents the common case that serves as a baseline in Table~\ref{tab_veri_typical_cases}. Case \#2 represents the use of a relatively risky strategy by reducing the battery safety threshold from 0.3 to 0.25. Indeed, in Table~\ref{tab_veri_typical_cases}, we see a decreased probability of mission success (from 1 to 0.91), whilst the expected mission time and number of recharges also significantly reduce, which is the benefit of taking more risk. Comparing case \#3 and \#1, given a fairly safe battery strategy (i.e. $\mathit{safe\_t}=0.3$), a more dynamic environment will significantly increase the mission time and number of recharges. Because the more dynamic the environment is, the more often the UAV decides to go back for recharges for battery safety reasons. Note, the probability of mission success remains (i.e. 1), since the battery strategy is conservative enough to guarantee a safe trip back to base in all possible circumstances. On the contrary, if we adopt a risky battery strategy in a more dynamic environment (\#4), then not only the expected mission time increases but also the probability of mission success decreases (cf. \#4 and \#2), because there are cases that the UAV does not reserve enough battery to fly back to base due to a sudden change of environments.

%%XZ: due to the page limits and potentially digression, leave out the results for later version.
%\begin{figure}[h!]
%	\centering
%	\includegraphics[width=1\textwidth]{fig_prob_suc_vs_cfull_vary_safet}
%	\caption{Probability of mission success, given different battery safety strategies in a normal working environment ($p\_wsp\_c=0.1$), as a function of the new battery capacity.}
%	\label{fig_prob_suc_vs_cfull_vary_safet}
%\end{figure}

%%XZ: due to the page limits and potentially digression, leave out the results for later version.
%\begin{figure}[h!]
%	\centering
%	\includegraphics[width=1\textwidth]{fig_prob_suc_vs_cfull_vary_wsp}
%	\caption{Probability of mission success, given different levels of environmental dynamics and a common battery safety strategy ($safe\_t=0.3$), as a function of the new battery capacity.}
%	\label{fig_prob_suc_vs_cfull_vary_wsp}
%\end{figure}

\subsection{Comparison of Models, Disregarding the Battery Features}

Most existing verification studies of autonomous robots, when considering energy constraints, formalise the energy component in a generic/simplified manner such that battery features are overlooked. In this section, we illustrate the difference between a simplified battery model and our relatively advanced model, considering the battery chemical features. 

\begin{figure}[ht]
	\centering
	\includegraphics[width=1\textwidth]{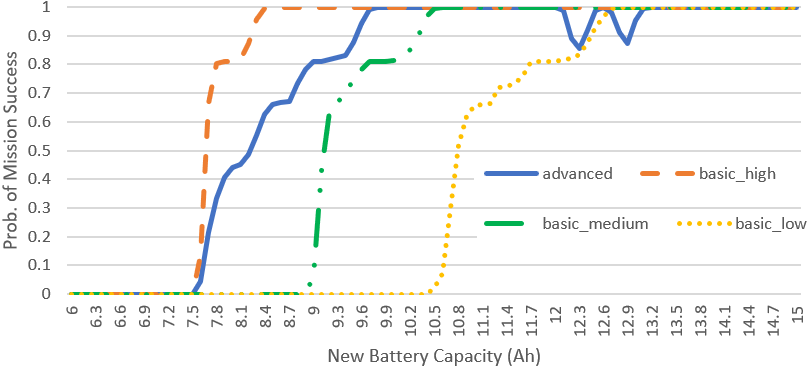}
	\caption{Probability of mission success, via different model assumptions on batteries, as a function of the new battery capacity.}
	\label{fig_prob_suc_vs_cfull_adv_basic}
\end{figure}

Fig.~\ref{fig_prob_suc_vs_cfull_adv_basic} shows the probability of mission success, via different models, as a function of the new battery's capacity (Ah). The solid curve labelled as ``advanced'' represents our proposed model considering the battery features (BA1 and BA3). The other curves represent the basic models without considering battery features, e.g. the ``basic\_high'' curve is the case when there is no capacity fading and the battery always works at a high level of voltage.

In Fig.~\ref{fig_prob_suc_vs_cfull_adv_basic}, we can observe that, for a given model, there is a required minimum new battery capacity to have non-zero probability to succeed. Indeed, the capacity should be at least enough for the inspection of the first wind turbine and a safe trip back. Since the battery consumption of each action is higher (and highest) when assuming that the battery is always working at a medium (and low) voltage level, such a required minimum capacity increases. Similarly, to guarantee a successful mission, ``basic\_high'' requires that the relatively smallest new battery capacity (around 8.4Ah) due to its obviously optimistic assumptions, i.e. no capacity fading and always working at a high voltage level.

%\xiaowei{would be useful to give a concrete path showing that the drone takes risk and fails. Also, a path for overly conservative path is useful.} XZ: ok. 

Note, although the ``advanced'' model is bounded by ``basic\_high'' and ``basic\_medium'', it is still dangerous to use such simplified bounds to do approximation, due to the observed ``dip'' on the ``advanced'' curve in the range of 12Ah--13Ah. That dip of probability of mission success happens because, when the new battery's capacity increases (but is still not big enough), the UAV may decide to take more risk to perform more actions in one trip (i.e. one discharge cycle), after which there might not be enough SOC left for a safe trip back in some edge cases (e.g. a degraded battery working at a low voltage in a high wind speed environment). To eliminate this phenomenon, the simplest way is to raise the battery safety threshold, which is confirmed by our extra experiments.

\begin{figure}[ht]
	\centering
	\includegraphics[width=1\textwidth]{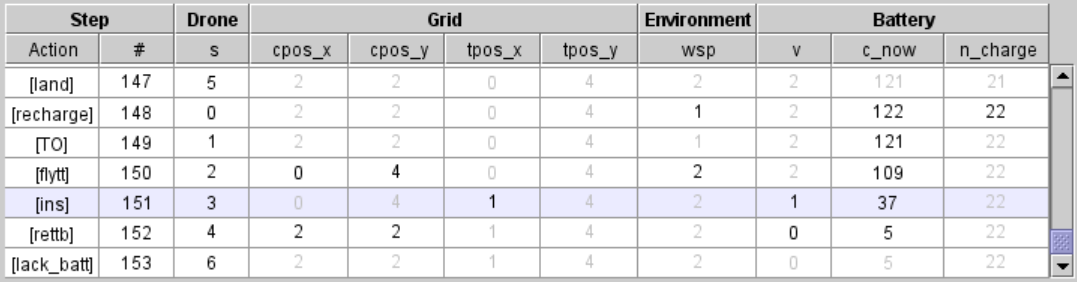}
	\caption{A fragment of a failed mission path generated by the PRISM simulator, which is an example of the ``dip'' in Fig.~\ref{fig_prob_suc_vs_cfull_adv_basic} with the new battery's capacity as 12.8Ah. Note, only key variables of the 4 modules are configured to be viewed here.}
	\label{fig_risky_path}
\end{figure}

An example path of a failed mission is presented in Fig.~\ref{fig_risky_path}, in which the new battery's capacity is 12.8Ah (thus within the ``dip'' range in Fig.~\ref{fig_prob_suc_vs_cfull_adv_basic}). After 22 recharges at step \#148, the UAV flies to the target cell [0,4] and the wind speed changes to high ($wsp=2$). At step \#150, the predicted SOC after the intended inspection is higher than the safety threshold of 0.3. So instead of returning to the base, the UAV continues the inspection in high-speed wind. Although after managing to return to base, the drone fails to land in a high wind speed and at the lower voltage level. Also, if we naively ignore the capacity fading and/or assuming the battery never works at a low voltage, then the UAV would land safely in this example. That's why we don't observe the ``dips'' on the curves of the basic models in Fig.~\ref{fig_prob_suc_vs_cfull_adv_basic}.

\begin{figure}[ht]
	\centering
	\includegraphics[width=1\textwidth]{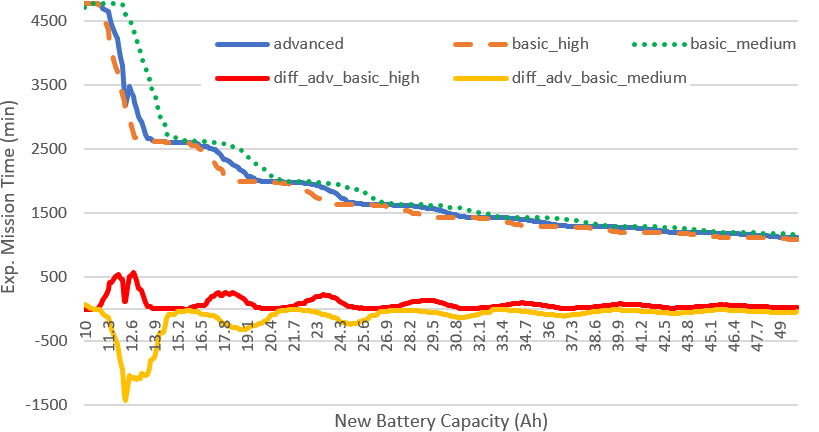}
	\caption{The upper graph curves show the expected mission time, for the specified battery model, as a function of the new battery capacity. The lower graph curves show the differences of the expected mission times from the specified models.}
	\label{fig_exp_miss_time_diff_VS_cfull_adv_basic}
\end{figure}

Fig.~\ref{fig_exp_miss_time_diff_VS_cfull_adv_basic} shows the expected mission time (upper graph curves) from the specified battery models and the differences (lower curves) between them. We observe that, in the practical range of the new battery's capacity (i.e. $<$13Ah, base on our survey), the basic models could give either too optimistic (500 minutes less) or over pessimistic (1500 minutes more) results. Such a variance of 1 to 3 working days will mislead wind farm maintenance activities and thus cause significant economic loss. Not surprisingly, as the new battery capacity tends to infinity (i.e. the battery is no longer a bottle neck of the given mission), the verification results of all models tends to the same value (as do the results in Fig.~\ref{fig_prob_suc_vs_cfull_adv_basic}).

\section{Related Work}
\label{sec_related_work}

How autonomous robots should be verified is a new challenging question \cite{fisher_verifying_2013,farrell_robotics_2018}, and it has received great attention in recent years, e.g. \cite{gerasimou_efficient_2014,norman_verification_2017,pathak_verification_2018,zhao_probabilistic_2019}. When considering energy constraints, the energy consumption is usually formalised in a linear way that being generic for both liquid-fuel and batteries. For instance, in the analysis of robot swarms \cite{konur_analysing_2012,liu_modeling_2010} the authors assume constant energy cost at each time step and a fixed capacity when obtaining energy from ``food''. Again, in the modelling of UAV missions \cite{giaquinta_strategy_2018,hoffmann_autonomous_2016}, a fixed battery capacity and constant battery consumption over time is assumed. In \cite{gerasimou_efficient_2014}, energy consumption of UUV sensors is modelled as a reward/cost for each state, which exhibits a linear behaviour over time. Indeed, such generic and simplified assumptions do not necessarily mean that they are unrealistic, whilst we believe more rigorous discussions and studies should be carried out prior to their adoption.

The study in \cite{boker_battery_2014} highlights the difference between real and ideal batteries, with a case study on controlling an energy-constrained robot. But it focuses on another battery feature -- ``recovery effect'' (e.g. a smart phone might shutdown due to an out-of-battery failure, but then become live again after an idle period).

Beyond the scope of robotics systems, battery behaviour does draw attention for verification. For instance, in \cite{wognsen_battery_aware_2014}, the battery of a satellite is described by the Kinetic Battery Model which is formalised as a timed automata to precisely model the discharge behaviour. However they leave out the capacity fading feature as future work. Similarly in \cite{ivanov_analytical_2018}, the Kinetic Battery Model is used for analysing wireless sensor protocols. For smartphones, \cite{espada_runtime_2015} uses runtime verification to check whether the actual battery consumption is within the expected limits that are derived from battery consumption profiles for smartphone apps.

\section{Discussions, Conclusions and Future Work}
\label{sec_conclusions}

In this paper, we formalise a UAV inspection mission of an offshore wind farm, and then do probabilistic model checking in PRISM to show (i) how the battery's non-linear features significantly affect the verification results in most practical cases; and (ii) how battery safety strategies, dynamic environments and battery features jointly affect the verification results.

Most existing formal verification studies of robots make simplified linear assumptions on energy consumption, which is indeed preferable in the case that the capacity is far beyond the total battery cost of the whole mission (i.e. when there is no recharges and the battery's working voltage is fairly stable due to a considerable SOC margin remaining at the end of the mission). In contrast, our work shows how such a simplification can significantly affect the verification results in the case that there are multiple recharges in the autonomous mission. Thus, we believe our work highlights this risk and calls for more rigorous discussions prior to any battery assumptions made in future formal verification of robots, especially when recharges are expected in the mission scenarios.

Moreover, we believe that battery PHM techniques should be explicitly integrated into the formal verification of robots. Although in this paper we use a hypothetical/generic battery PHM technique to provide the parameters used in the \textit{Battery} module, it is clear how PHM can aid the rigorous modelling of battery for formal verification. For now, both the battery PHM experiments and formal verification are assumed to be carried out in the lab, i.e. prior to the mission. To improve the accuracy, an appealing idea is to integrate both at runtime, since there is a trend of doing online battery PHM based on real-time readings from the sensors deployed on the battery, e.g. \cite{barre_real_time_2014,zhang_online_2018}. Whilst there will be a scalability issue if running both online battery PHM and formal verification algorithms at runtime. A compromised solution, in our example, is to invoke the formal verification and PHM during the recharging at the base (where substantial computing resources can be used) with newly collected log-data from recent flights. Thus the verification result will be updated with the up-to-date data. We plan to implement this solution in our future work.

Apart from highlighting the need of integrating battery PHM techniques, this work only serves as a first approximation\footnote{It is a first approximation in the sense of, e.g. the simplification of two levels of wind speed and the round estimations of battery consumption in Tab.~\ref{tab_battery_capacity_use_actions_diff_voltage}.} of the verification of the residential drone inspection mission. More rigorous verification/planning of the mission is needed in future, e.g. by gradually refining the fundamental PRISM model based on observations from various sources of data \cite{paterson_observatio_2018,paterson_using_2019}.

In summary, our main contributions are:
\begin{itemize}
    \item We formalise a UAV inspection mission on a wind farm based on a real industry survey project, which can be reused and extended as an exemplar for future research of similar UAV missions.
    \item We do a sequence of what-if calculations, via probabilistic model checking, to show (i) the importance of considering non-linear battery features in formal verification of autonomous robots; and (ii) how such battery features, together with the dynamic environments and battery safety strategy, jointly affect the verification results.
    \item We discuss the need of explicitly integrating battery PHM techniques into formal verification of robots, and propose a potential solution which forms important future work.
\end{itemize}

%\begin{theorem}
%This is a sample theorem. The run-in heading is set in bold, while
%the following text appears in italics. Definitions, lemmas,
%propositions, and corollaries are styled the same way.
%\end{theorem}
%
% the environments 'definition', 'lemma', 'proposition', 'corollary',
% 'remark', and 'example' are defined in the LLNCS documentclass as well.
%

%
% ---- Bibliography ----
%
% BibTeX users should specify bibliography style 'splncs04'.
% References will then be sorted and formatted in the correct style.
%
 \bibliographystyle{splncs04}
 \bibliography{ref}
%
%\begin{thebibliography}{8}
%\bibitem{ref_article1}
%Author, F.: Article title. Journal \textbf{2}(5), 99--110 (2016)
%
%\bibitem{ref_lncs1}
%Author, F., Author, S.: Title of a proceedings paper. In: Editor,
%F., Editor, S. (eds.) CONFERENCE 2016, LNCS, vol. 9999, pp. 1--13.
%Springer, Heidelberg (2016). \doi{10.10007/1234567890}
%
%\end{thebibliography}
\end{document}